# Unifying Local and Global Change Detection in Dynamic Networks

**Wenzhe Li, Dong Guo, Greg Ver Steeg, Aram Galstyan**
University of Southern California, Information Sciences Institute
Marina del Rey, CA 90292
{wenzheli,dongguo,gregv,galstyan}@isu.edu

## Abstract

Many real-world networks are complex dynamical systems, where both local (e.g., changing node attributes) and global (e.g., changing network topology) processes unfold over time. Local dynamics may provoke global changes in the network, and the ability to detect such effects could have profound implications for a number of real-world problems. Most existing techniques focus individually on either local or global aspects of the problem or treat the two in isolation from each other. In this paper we propose a novel network model that simultaneously accounts for both local and global dynamics. To the best of our knowledge, this is the first attempt at modeling and detecting local and global change points on dynamic networks via a unified generative framework. Our model is built upon the popular mixed membership stochastic blockmodels (MMSB) with sparse co-evolving patterns. We derive an efficient stochastic gradient Langevin dynamics (SGLD) sampler for our proposed model, which allows it to scale to potentially very large networks. Finally, we validate our model on both synthetic and real-world data and demonstrate its superiority over several baselines.

## Introduction

Networks are one of the most popular tools for modeling interactions among objects or people. Diverse problems can be formulated in terms of networks; these include social network mining (Zhu et al. 2016; Liben-Nowell and Kleinberg 2007), analysis of protein-protein interactions (Shen et al. 2007; Bader and Hogue 2003; Brohee and Van Helden 2006), and fraud detection (Chiu et al. 2011; Šubelj, Furlan, and Bajec 2011). One major challenge of network modeling is to incorporate realistic dynamics. In fact, most common networks change their structure over time. As examples, in social networks friendship links are sporadically added or removed, and in communication networks, the frequency and volume of transmitted data changes continuously over time. Thus, it is important to have efficient computational models that can be used to adequately describe network dynamics.

One problem in particular that has attracted significant recent interest is detection of important changes in dynamic networks. Such changes can occur both at the node level and for the overall structure. Consider the co-authorship network where a node represents a person and dynamic links result from co-authoring a paper at a specific time. The research interest of a person may change over time, i.e., from Bayesian inference to deep learning, which can be regarded as a local change. In addition, the overall collaboration patterns might change as well. Indeed, while scientific collaboration traditionally happened between researchers with similar interests, recent growth of interdisciplinary subjects has encouraged and/or necessitated collaboration between researchers with more diverse backgrounds, which can be regarded as a significant global change to the network. Furthermore, these two types of changes are deeply intertwined, prompting the following question: *How can we detect both local and global changes at the same time*? Despite considerable previous work on detecting changes, to the best of our knowledge, none of the previous models consider both types of changes in a unified framework. For instance, the model in (Peel and Clauset 2015) focuses on detecting global changes in network structure while discarding local dynamics. Similarly, the work in (Raghavan et al. 2014) considers a model where nodes evolve locally according to a Markovian dynamics, but the global connectivity patterns remain unchanged.

Here we propose a computational approach that jointly incorporates both local and global aspects of network dynamics. Our model is based on mixed membership stochastic blockmodels (MMSB) (Airoldi et al. 2008), where each node is characterized by a normalized membership vector over finite number of *roles*, and interactions are role-dependent. MMSB differs from popular stochastic blockmodels (SBM) (Karrer and Newman 2011; Wang and Wong 1987) in that it allows each node to have multiple roles. This is especially useful when our goal is to understand the transition patterns of membership for each node; i.e., we can observe how each node evolves over time by looking at its mixed membership vectors. In the context of MMSB, the global changes can be encoded in the so called affinity (or compatibility) matrix.

To account for local dynamics, we use a selection and influence mechanism proposed in (Cho, Steeg, and Galstyan 2011). In contrast to (Cho, Steeg, and Galstyan 2011), however, we allow the affinity matrix to change over time. This is the first essential step for enabling the detection of global changes. Furthermore, we introduce sparse weighting factors to control both the transitions for local membership vectors and the fraction of nodes that experience considerable changes between two consecutive time points. The sparse weighting factors play an important role in simultaneously detecting the local and global changes. In the experimental results, we will show the importance of such changes by comparing the newly proposed model to one without any sparsity constraints.

Throughout the paper we are dealing with Bayesian graphical models. Similar to many other Bayesian models such as latent Dirichlet allocation (LDA) (Blei, Ng, and Jordan 2003), Bayesian neural networks (Hernández-Lobato and Adams 2015), etc., the exact inference for MMSB-type models is also impossible. In that case, we have to rely on some forms of approximate inference techniques such as Markov Chain Monte Carlo (MCMC) method. However, classical MCMC is too slow to apply to large data. More recently, a new scheme for MCMC samplers has been proposed; it is called SGLD (Stochastic Gradient Langevin Dynamics) (Welling and Teh 2011), and can be regarded as the stochastic variant of Langevin dynamics.[1] Under the framework of SGLD, we can generate samples by only looking at a mini-batch of data. In fact, this significantly improves learning speed, especially for very large data. Previous work (Bhadury et al. 2016; Li et al. 2016; El-Helw et al. 2016) demonstrates that the sampling scheme based on SGLD performs better than other approximate techniques such as stochastic variational inference.

The contributions of this paper can be summarized as:

- We propose a novel model for evolving networks that incorporates both local and global dynamics. To the best of our knowledge, this is the first unified generative framework for effectively detecting both local and global changes in dynamic networks at the same time.
- We provide an efficient SGLD inference procedure for our model. While SGLD has been used for static networks, here we provide a non-trivial extension of the method for co-evolving network model.
- We conduct comprehensive experiments on both synthetic and real-world data sets, and we show that our new model significantly outperforms other baselines in terms of both quantitative and qualitative measures.

The rest of the paper is organized as follows: First, we give an overview of related work on dynamic network and change detection, followed by preliminaries on mixed membership stochastic blockmodels. Then, we provide detailed description of our proposed model and its scalable sampling procedure. Finally, experiments on both synthetic and real-world data are presented.

## Related Work

**Dynamic Networks.** There are several works on dynamic networks that are related to ours. (Fu, Song, and Xing 2009) proposed dynamic mixed membership stochastic blockmodels for evolving networks. The method they use to model the dynamic transition is to incorporate changing priors over membership vector for each node, which is built upon the well-known state space model (SSM) (Kitagawa 1996; Roesser 1975). Since all the subsequent node membership at a particular time point $t$ is drawn from the same prior distribution, the learned model potentially leads to lack of diversity for nodes, or the flexibility of describing the underlying dynamics. (Cho, Steeg, and Galstyan 2011) proposed the co-evolving MMSB which incorporates selection and influence mechanisms. One important assumption about the model is that the evolution of a node is driven by the interactions between nodes through a parameterized influence mechanism. Since the evolution of one node is relatively independent of others, the model has better flexibility for capturing local changes, but with the added cost of introducing more parameters. Although both approaches are designed to model the dynamics of the network, they lack the capability for capturing both local and global changes. As shown in (Cho, Steeg, and Galstyan 2011; Fu, Song, and Xing 2009), all the experiments are conducted by fixing the global affinity matrix.

**Change Detection in Networks.** (Peel and Clauset 2015) used a parametric approach to change detection in networks by assuming a generalized hierarchical random graph model. (Ridder, Vandermarliere, and Ryckebusch 2016) extended the previous work to include stochastic blockmodels. Both approaches are optimized for detecting large-scale structural changes in the network, but uncapable of capturing localized changes in the network dynamics. (Barnett and Onnela 2016) proposed a change detection method that makes minimal distributional assumptions. However, their approach is tailored to correlational networks. (Cadena, Vullikanti, and Aggarwal 2016) developed event detection method that works for both signed and unsigned networks, but is limited to specific type of structural change – emergence of anomalously dense subgraphs. (Kolar and Xing 2012) uses fused-type penalty combined with an l1 penalty to estimate the jumps from dynamic networks, but mainly used to model the structural changes. Note finally that a related line of work has focused on detecting anomalies in dynamic networks; see (Akoglu, Tong, and Koutra 2015) for a recent survey.

## Preliminaries

Consider a sequence of networks $\{\mathcal{G}^t\}_{t=1}^T$, where $\mathcal{G}^t = (\mathcal{V}, E^t)$ is the network snapshot observed at time $t$. We set the size of $|\mathcal{V}| = N$ and allow the network connectivity to change over time.

The goal of our model is to simultaneously capture both global and local transition patterns from data. In our model we use a temporal sequence of the so-called *affinity matrix* to capture global changes and a sequence of *mixed membership vectors* to capture the local changes for each node. Mixed membership stochastic blockmodels (MMSB) (Airoldi et al. 2008) model the network based on a given affinity matrix and membership vectors.

To build the dynamic model, we use a selection and influence mechanism (Cho, Steeg, and Galstyan 2011) so that, at each time point $t$, the mixed membership for each node is influenced by the membership of its neighbors as well as itself at time $t-1$. The strength of impact of neighbors can be controlled by model parameters, which we will show in the next section. In addition, we make an important assumption that *between every two consecutive time points, the portion of nodes which experience considerable change should be small*. In fact, in the real-world setting, the membership/roles remain stable for a majority of the nodes or entities. We introduce a sparse prior for coping with this constraint, which leads to rather significantly different results compared to the case when there is no such constraint.

[1] Langevin dynamics is a type of MCMC sampler

## Mixed Membership Stochastic Blockmodels (MMSB)

Formally, for each node $p \in \mathcal{V}$ has a $K$ dimension probability distribution $\pi_p$ of participating in the $K$ members of the community set $\mathcal{K}$. We also refer to $\pi_p$ as the (normalized)[2] *mixed membership vector* whose entries sum to 1.

For each pair of nodes $p$ and $q$, we have a binary scalar observation $Y_{p,q} \in \{0, 1\}$. If $p$ and $q$ are connected, $Y_{p,q}$ becomes 1, 0 otherwise. Under the stochastic framework of MMSB, $Y_{p,q}$ is stochastically generated based on the mixed membership vectors $\pi_p, \pi_q$ and affinity matrix $\mathcal{B}$. The affinity matrix controls how likely two nodes in differing (or the same) communities are to connect to each other.

As a reminder of the generative process, we first draw the interaction indicators $z_{p\rightarrow q}, z_{p\leftarrow q} \in \{1, , K\}$, each of which is drawn from mixed membership vectors $\pi_p, \pi_q$, respectively. Then we draw a Bernoulli random variable from the particular entry of $\mathcal{B}$, where the position of entry is decided by interaction indicators $z_{p\rightarrow q}, z_{p\leftarrow q}$. i.e. $Y^t_{p,q} \sim \mathrm{Ber}(\mathcal{B}_{kl})$ where $z_{p\rightarrow q} = k, z_{p\leftarrow q} = l$. The detailed generative process of MMSB is then given by:

1. For each node $p$, draw node membership $\pi_p \sim \mathrm{Dir}(\alpha)$
2. For each pair of nodes $p$ and $q$,
   (a) Draw interaction indicator $z_{p\rightarrow q} \sim \mathrm{Mul}(\pi_p)$
   (b) Draw interaction indicator $z_{p\leftarrow q} \sim \mathrm{Mul}(\pi_q)$
   (c) Draw link $Y_{pq} \sim \mathrm{Ber}(\gamma)$, where $\gamma = B_{kl}$ such that $z_{p\rightarrow q} = k, z_{p\leftarrow q} = l$

where Ber(.) denotes drawing from Bernoulli distribution, Mul(.) from Multinomial distribution, Dir(.) from Dirichelt distribution.

The affinity matrix $\mathcal{B}$ plays an important role for the network generation process. If we set $\mathcal{B}$ as a diagonal matrix, only nodes with the same community are connected; If we set $\mathcal{B}$ as a full matrix, then each pair of communities has some chance to be connected. In addition, the block-diagonal matrix $\mathcal{B}$ can also be used to generate the particular structure of a graph. Thus, the flexibility of choosing different patterns of $\mathcal{B}$ provides strong expressive power for dealing with corresponding network structures.

Variational inference (Airoldi et al. 2008; Blei, Ng, and Jordan 2003) and MCMC sampling (Ishwaran and James 2001; **?**) are two popular ways to infer posteriors from Bayesian graphical models such as MMSB. Both of them use the full batch of data at each iteration, which potentially precludes the use of such models with larger data or streaming data. More recently, a stochastic version of variational inference (Gopalan et al. 2012; Patterson and Teh 2013) and MCMC sampler (Ahn et al. 2014; Welling and Teh 2011; Ahn et al. 2014; Li et al. 2016; El-Helw et al. 2016) were developed, where we can effectively use mini-batch updates for each iteration.

[2]Later, we will introduce another un-normalized mixed membership vector $\mu_p$ for the dynamic model.

## Sparse Co-evolving MMSB (SC-MMSB)

In the dynamic setting, it is possible that the mixed membership vectors for nodes and the affinity matrix $\mathcal{B}$ can evolve over time, and our goal is to seamlessly detect such local and global changes in a unified framework. As we mentioned earlier, we use selection and influence mechanisms to build our model. In particular, for each node $p \in \mathcal{V}$ at time $t$, we first define (unnormalized) membership vector $\boldsymbol{\mu}^t_p$, which can map to simplex ($\boldsymbol{\pi}^t_p$) via the following logistic transformation function $g(.)$:

$$\pi^t_{p,k} = g(\mu^t_{p,k}) = \exp(\mu^t_{p,k} - C(\boldsymbol{\mu}^t_p)),\ \ \forall k = 1, ..., K$$

Here, $C(\boldsymbol{\mu}^t_p)$ is a normalization constant (i.e., the log partition function) such that $C(\boldsymbol{\mu}^t_p) = \log(\sum_{k=1}^K \exp(\mu^t_{p,k}))$. Due to such normalizability constraint, $\boldsymbol{\pi}^t_p$ only has $K - 1$ degrees of freedom.

We next define the evolution of the mixed membership vector for node $p$. At each time point $t$, we draw the node membership from the following Gaussian distribution:

$$\boldsymbol{\mu}^t_p \sim \mathcal{N}(f(\boldsymbol{\mu}^{t-1}_p, \boldsymbol{c}^{t-1}_{m(p)}), \boldsymbol{\Sigma}) \qquad (1)$$

As we can see, the new membership vector mainly depends on three components: its own membership $\boldsymbol{\mu}^{t-1}_p$ at the previous time point, the information gathered from its neighbors at time $t-1$ denoted as $\boldsymbol{c}^{t-1}_{m(p)}$, as well as the function $f(.)$ which controls the combination of these two sources of information. $\boldsymbol{\Sigma}$ is the covariance matrix, and we set it as diagonal in this paper. $\eta_k$ is the variance for dimension $k$ and we estimate this quantity from data.

The flexibility of such a modeling approach comes from different choices of function $f(.)$ and the computation of $\boldsymbol{c}^{t-1}_{m(p)}$. First, the function $f(.)$ can be set as either linear or nonlinear form. In this paper we assume it follows the weighted average scheme as in Eqn 2. $\beta^t_p$ is the weighting parameter to control the trade off between two major quantities. In a more complicated scenario (i.e., deep learning), the function itself can contain a set of parameters to learn some forms of embedding in the latent space. Second, there are many ways to integrate the information from neighbor nodes. For instance, whether to include indirect neighbors in addition to direct ones, etc. One could also randomize and choose a subset of neighbors (Grover and Leskovec 2016). Choosing the simplest option, we compute the value of $\boldsymbol{c}^{t-1}_{m(p)}$ with the average of its (1st degree) direct neighbors at time $t-1$. We leave exploration of more complex choices of function $f(.)$ and neighborhood selection schemes to future work. The full form of function $f(.)$ can be written as:

$$f(\boldsymbol{\mu}^{t-1}_p, \boldsymbol{c}^{t-1}_{m(p)}) = (1 - \beta^t_p)\boldsymbol{\mu}^{t-1}_p + \beta^t_p \boldsymbol{c}^{t-1}_{m(p)} \qquad (2)$$

then the term $\boldsymbol{c}_{m(p)}^{t-1}$ can be computed:

$$\boldsymbol{c}_{m(p)}^{t-1} = \frac{1}{|m(p)^t|} \sum_{q \in m(p)^t} \boldsymbol{\mu}_q^{t-1} \tag{3}$$

$m(p)^t$ denotes the set of neighbor nodes of $p$ at time $t$, and $|m(p)^t|$ is the size of this set.

The value of $\beta_p^t$ ranges from $[0, 1]$ controlling the importance of information gathered from neighbors on our current estimate of the mixed membership. If $\beta_p^t$ approaches to 0, then the node membership remains stable and the fluctuation is only managed by the covariance matrix $\boldsymbol{\Sigma}$. In practice, it makes sense to assume that there are only a small fraction of nodes that experience considerable membership changes between two consecutive time points, while others remain stable. For this purpose, we put a sparsity prior on each $\beta_p^t$, namely the Laplace prior. Because the value of $\beta_p^t$ must lie between $[0, 1]$, we use a truncated Laplace prior instead. By changing the parameter of the Laplace prior, we can manage the fraction of nodes that we allow to undergo considerable change. The sparsity constraint is very useful for disentangling the local and global changes for a dynamic model.

The transition of the affinity matrix $\mathcal{B}^t$ over time points remains the same as in (Cho, Steeg, and Galstyan 2011). Similarly, we first define un-normalized affinity matrix $\boldsymbol{\phi}^t$. The mapping from $\boldsymbol{\phi}^t$ to $\mathcal{B}$ is obtained by applying the element-wise sigmoid function $\sigma(.)$ on $\boldsymbol{\phi}^t$, $\mathcal{B}_{ij} = \sigma(\phi_{ij})$. At each time point $t$, we assume the value of $\phi_{ij}^t$ follows the Gaussian distribution:

$$\phi_{ij}^t \sim \mathcal{N}(\phi_{ij}^{t-1}, \gamma^2) \tag{4}$$

where $\gamma^2$ is the variance that is constant over all dimensions.

Given global affinity matrix $\mathcal{B}$ and mixed membership for each node, the stochastic process of network generation remains the same as static MMSB. Combining these ingredients leads to the generative process shown below. Figure 1 depicts the graphical representation of our model.

1. **State Space Model for Local Membership**
   - For each node $p$ at time $t = 0$, $\boldsymbol{\mu}_p^0 \sim \mathcal{N}(\boldsymbol{\alpha}^0, \mathbf{A})$
   - For each node $p$ at time $t > 0$, $\boldsymbol{\mu}_p^t \sim \mathcal{N}(f(\boldsymbol{\mu}_p^{t-1}, \boldsymbol{c}_{m(p)}^{t-1}), \boldsymbol{\Sigma})$
   - For each node $p$ at time $t$, $\beta_p^t \sim$ Laplace$(b)$
2. **State Space Model for Global Affinity Matrix**
   - For each community pair $k, l$ at time 0, $\phi_{k,l}^0 \sim \mathcal{N}(\iota, \sigma^2)$, where $k, l \in \{1, .., K\}$
   - For each community pair $k, l$ at time $t > 0$, $\phi_{k,l}^t \sim \mathcal{N}(\phi_{k,l}^{t-1}, \gamma^2)$
3. **Network Generation Model for Time Point** $t$
   - For each pair of node $(p, q)$:
     - $z_{p\rightarrow q}^t \sim$ Mul$(g(\boldsymbol{\mu}_p^t))$
     - $z_{p\leftarrow q}^t \sim$ Mul$(g(\boldsymbol{\mu}_q^t))$
     - $Y_t(p, q) \sim$ Ber$((1-\rho)\sigma(\boldsymbol{\phi}_{z_{p\rightarrow q}^t, z_{p\leftarrow q}^t}))$

where $\rho$ implies the sparsity of network connections.

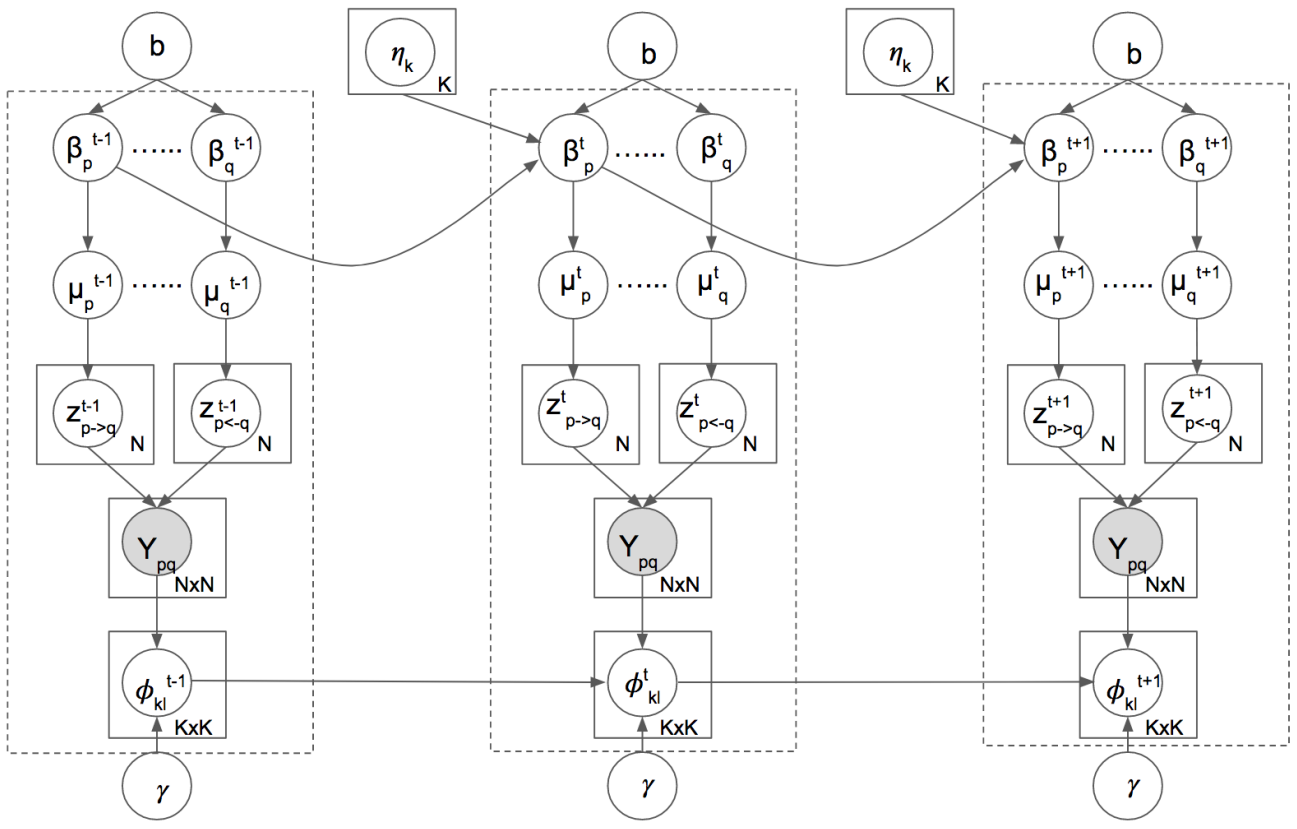

Figure 1: A graphical representation of our proposed model (SC-MMSB). The arrows encode the dependency structure.

## Inference

Like many other Bayesian graphical models, exact inference for SC-MMSB is intractable. Thus, approximate inference techniques such as MCMC should be applied. However, classical MCMC is hard to scale up for large data because it is computationally intensive. Recently, the state-of-art stochastic MCMC scheme, SGLD (Stochastic Gradient Langevin Dynamics) is proposed (Welling and Teh 2011) which generates samples by only looking at mini-batch data. This significantly reduces the cost for generating samples, and shown to be more effective than another competing method called stochastic variational inference (Bhadury et al. 2016; Li et al. 2016; El-Helw et al. 2016).

SGLD sampler is built upon the well-known technique, stochastic optimization, and adds some Gaussian noise to each update so that it can generate samples from the true posterior and avoid collapsing to a MAP/MLE solution. Let $p(\boldsymbol{\theta}|\{X_n\}) \propto p(\boldsymbol{\theta}) \prod_n p(x_n|\boldsymbol{\theta})$ be an arbitrary generative model with prior $p(\boldsymbol{\theta})$ and likelihood $p(x|\boldsymbol{\theta})$. The SGLD parameter update for $\boldsymbol{\theta}$ at the $i$th iteration is:

$$\triangle\boldsymbol{\theta}_i = \frac{\epsilon_i}{2}\left(\triangledown \log p(\boldsymbol{\theta}) + \frac{N}{M}\sum_{n=1}^{M} \triangledown \log p(x_n|\boldsymbol{\theta})\right) + \xi_i$$

$$\xi_i \sim \mathcal{N}(\xi_i|0, \epsilon_i)$$

where $N$ is the number of data points in the training set, and $M$ is a mini-batch sampled from the full batch. In general, we use $\epsilon_i = a \times (b + i)^{-c}$ for the noise term. In our inference procedure, we estimate the latent variables $\boldsymbol{\mu}_p^t, \phi_{kl}^t, z_{p\rightarrow}^t, z_{p\leftarrow q}^t$ using MCMC sampler, and the model parameters $\eta_k, \gamma, \beta_p^t$ with closed-form solution.

**Sampling $\boldsymbol{\mu}_p^t$** It is easy to see from Figure 1 that the Markov Blanket (MB) contains the terms $\boldsymbol{\mu}_p^{t+1}, \boldsymbol{\mu}_p^{t-1}$ and $\boldsymbol{\mu}_q^{t+1}$ for $q \in m(p)^t$. Then we can write down the probability of $\mu_{p,k}^t$ given all the variables within the Markov Blanket

as:

$$p(\mu^t_{p,k}|MB(\mu^t_{p,k}) \propto \mathcal{N}(\mu^t_{p,k}|f(\mu^{t-1}_{p,k}, c^{t-1}_{m(p),k}), \eta^2_k) \times \mathcal{N}(\mu^{t+1}_{p,k}|f(\mu^t_{p,k}, c^t_{m(p),k}), \eta^2_k) \times \prod_{q\in m(p)^t} \mathcal{N}(\mu^{t+1}_{q,k}|f(\mu^t_{q,k}, c^t_{m(q),k}), \eta^2_k) \times \prod_{q\neq p} \text{Mul}(z^t_{p\to q}|g(\mu^t_{p,k})) \quad (5)$$

The first three terms are the product of Gaussian, which is also a Gaussian distribution, playing as a prior for $\mu^t_{p,k}$. By using the "completing the square" trick, we can get the new Gaussian form for $\mu^t_{p,k}$.

$$\mu^t_{p,k} \sim \mathcal{N}(\frac{B}{A}, \frac{\eta^2_k}{A}) \quad (6)$$

$$A = 1 + (1-\beta^{t+1}_p)^2 + \sum_{q\in m(p)^t} \beta^{t+1\,2}_q \frac{1}{|m(q)^t|^2}$$

$$B = (1-\beta^t_p)(\mu^{t-1}_{p,k} + \mu^{t+1}_{p,k} - \beta^{t+1}_p c^t_{m(p),k}) + \beta^t_p c^{t-1}_{m(p),k} + \sum_{q\in m(p)^t} \beta^{t+1}_q \frac{1}{|m(q)^t|}(\mu^{t+1}_{q,k} - (1-\beta^{t+1}_q)\mu^t_{q,k} - \beta^{t+1}_q \frac{1}{|m(q)^t|} \sum_{r\neq p, r\in m(q)^t} \mu^t_{r,k})$$

The evaluation of $B$ takes $O(|m(q)^t||m(p)^t|)$ steps. Thus, the time complexity for sampling each $\boldsymbol{\mu}^t_p$ takes $O(K|m(q)^t||m(p)^t|)$, where $K$ is the number of communities.

For the likelihood term $\prod_{q\neq p}\text{Mul}(z^t_{p\to q}|g(\mu^t_{p,k}))$, after taking the logarithm, we have

$$\log \mathcal{L} = \sum_{i=1}^{K} C^t_{p,i} \log g(\mu^t_p)_i = C^t_{p,k}\mu^t_{p,k} - |m(p,t)| \log \sum_j \exp(\mu^t_{p,j}) \quad (7)$$

where $C^t_{p,k}$ denotes the number of $z_{p\to q}$ for $p$ assigned to community $k$ for all $q \in \mathcal{V}$. Finally, we get the SGLD update equation for $\mu^t_{p,k}$ as follows:

$$\mu^{t\,*}_{p,k} = \mu^t_{p,k} + \frac{\epsilon_i}{2}\left(-\frac{A}{\eta^2_k}(\mu^t_{p,k} - \frac{B}{A}) + C^t_{p,k} - |m(p)^t|g(\mu^t_{p,k})\right) + \xi_i \quad (8)$$

where $\xi_i \sim \mathcal{N}(0, \epsilon_i)$

**Sampling $z^t_{p\to q}, z^t_{p\leftarrow q}$** Given $\boldsymbol{\phi}$ and $\boldsymbol{\mu}$, the local latent variable $z$ is conditionally independent of the rest of the parameters in the model. When $y^t_{p,q} = 1$, the sampling for $z^t_{p\to q}$ becomes:

$$p(z^t_{p\to q} = k|.) \propto \exp(\mu^t_{p,k})\exp(\phi^t_{k,z^t_{p\leftarrow q}})$$

$$p(z^t_{p\leftarrow q} = k|.) \propto \exp(\mu^t_{q,k})\exp(\phi^t_{k,z^t_{p\to q}})$$

On the other hand, when $y = 0$, the derivation becomes:

$$p(z_{p\to q} = k) \propto g(\mu^t_p)_k[1 - (1-\rho)\sigma(\phi_{k,Z^t_{q\to p}})]$$

$$p(z_{p\leftarrow q} = k) \propto g(\mu^t_q)_k[1 - (1-\rho)\sigma(\phi_{k,z^t_{p\to q}})]$$

Because we need to compute the normalization term, it takes $O(K)$ time to get each sample for $z^t_{p\to q}$ and $z^t_{p\leftarrow q}$. Theoretically, a faster sampling approach (Bhadury et al. 2016) can be applied to reduce the complexity from $O(K)$ to an amortized cost of $O(1)$.

**Sampling $\phi^t_{k,l}$** As we see from Figure 1, the Markov Blanket for $\phi^t_{k,l}$ contains the terms $\phi^{t-1}_{k,l}, \phi^{t+1}_{k,l}, \mathbf{Y}^t, \mathbf{Z}^t$, where we use capital letters to represent the set of variables. The probability of $\phi^t_{kl}$ given all the variables within the Markov Blanket is:

$$p(\phi^t_{k,l}|\text{MB}(\phi^t_{k,l})) = \mathcal{N}(\phi^t_{k,l}|\phi^{t-1}_{k,l}, \gamma^2) \mathcal{N}(\phi^{t+1}_{k,l}|\phi^t_{k,l}, \gamma^2) \prod_{p=1}^{N}\prod_{q=p+1}^{N} P(Y_(p,q)^t|\mathcal{B}_{z_{p\to q}, z_{q\to p}})$$

The first two terms are Gaussian, so their product is also Gaussian which serves as the prior for $\phi^t_{k,l}$, The last one is the likelihood term which is the product of multinomial distribution. This formulation again naturally fits to the SGLD update framework. By incorporating two Gaussian terms, we can write the gradient w.r.t $\phi_{i,j}, i \leq j$ as:

$$\frac{\partial L}{\partial \phi_{i,j}} = -\frac{2}{\gamma^2}(\phi^t_{i,j} - \frac{\phi^{t-1}_{i,j} + \phi^{t+1}_{i,j}}{2}) + \sum_{k=1}^{K}\sum_{k'=k}^{K} \frac{C^1_{k,k'} - (1-\rho)C^{0+1}_{k,k'}\sigma(\phi^t_{k,k'})}{\sigma(\phi^t_{k,k'})[1-(1-\rho)\sigma(\phi^t_{k,k'})]} \sigma(\phi^t_{k,k'})[\delta(i=k, j=k') - \sigma(\phi^t_{i,j})] \quad (9)$$

where $C^t_{k,k'}$ counts the number of edges which satisfy $z^t_{p\to q} = k, z^t_{p\leftarrow q} = k'$. In the equation, we use two terms $C^1_{k,k'}, C^{0+1}_{k,k'}$ for ease of understanding. $C^1_{k,k'}$ denotes how many of the linked edges satisfy $z^t_{p\to q} = k, z^t_{p\leftarrow q} = k'$ while $C^{0+1}_{k,k'}$ denotes how many of both linked and non-linked edges satisfy those constraints.

**Learning Model Parameters** For $\beta^t_p$, $\eta_k$, and $\gamma$ there exists closed-form updates. For $\beta^t_p$, there are two free parameters $\mu, b$ where $b$ controls the shape of the distribution [3]. By using sub-gradient descent, we get the update equation as follows:

$$\beta^t_p = \frac{A - 1/b}{B} \text{ if } A > \frac{1}{b}, \text{ 0 otherwise}$$

$$A = -\sum_k (\mu^t_{p,k} - \mu^{t-1}_{p,k})(\mu^{t-1}_{p,k} - c^{t-1}_{m(p),k})/\eta^2_k$$

$$B = \sum_k (\mu^{t-1}_{p,k} - c^{t-1}_{m(p),k})^2/\eta^2_k$$

[3] In this paper, we fix the value of the first parameter

By changing the parameter $b$, we get the desired level of sparsity. Similar updates are obtained for parameter $\eta_k$, $\gamma$.

## Experiments

We first use synthetic data to demonstrate the effectiveness of our model, then apply it to two real-world data sets.

We introduce three baseline models, *dMMSB*, *cMMSB* and *GHRG-GLR*. *dMMSB* is proposed in (Fu, Song, and Xing 2009). The original model does not include the dynamics of the affinity matrix; they use a fixed affinity matrix in all of their experiments. In order to conduct the experiment in a similar setting, we slightly modify their model so that we force the affinity matrix to evolve over time. Also, in our implementation, we make a minor change to the transition of membership prior and let it follow a diagonal form. *cMMSB* is similar to our proposed new model except that there is no sparse prior on $\beta_p^t$. Last, *GHRG-GLR* is proposed in (Peel and Clauset 2015) which is specifically designed for global change detection while ignoring the local changes.

### Experiments on synthetic data

We generate three different variants of synthetic data sets, `synthetic 1`, `synthetic 2` and `synthetic 3`. For the first two, we set $K = 3, N = 30, T = 9$, and for the last one we set $K = 3, N = 30, T = 12$. We initially make the first set of 10 nodes belong to community 1, the next set of 10 nodes to community 2, and the remaining nodes to community 3. Once we have the affinity matrix and the local membership vectors for nodes, we can stochastically generate the network connections.

The `synthetic 1` data only simulates the global changes: for $t = 1, 2, 3$, the affinity matrix has major values at positions $(0, 0), (0, 1), (1, 0), (1, 1)$; for $t = 4, 5, 6$ the major values are at positions $(1, 1), (1, 2), (2, 1), (2, 2)$; and for the remaining time points, the major values are in its diagonal. In contrast to `synthetic 1`, `synthetic 2` data only simulates the local changes while making the affinity matrix a fixed diagonal form. We manually inject the abrupt changes into nodes $13 - 17$ at time point 5. Last, for `synthetic 3` data, we introduce both local and global changes. We simply combine the changes made from `synthetic 1` and `synthetic 2`.

Figure 2 shows the quantitative evaluation of three different models on the synthetic data. The performance is measured by both perplexity and Akaike information criterion (AIC) scores (Sakamoto, Ishiguro, and Kitagawa 1986), which are popular measure for quality of learned models. In addition to the quality of data fitting, AIC also takes the model complexity into account. From the results, we can easily see that our new model clearly outperforms two other competing baselines [4]. For this experiment, *dMMSB* performs worst, while the sparse *SC-MMSB* generally matches or out-performs the non-sparse *cMMSB*.

In Figure 3 we show the learned mixed membership vectors for `synthetic 2` data [5], where each color denotes a single community. Since the ground truth only includes the change of local membership for particular nodes, we expect that a good model should be able to recover the local changes while letting the learned affinity matrix remain stable over time. From the results of SC-MMSB, we can see that the nodes $13, 14, 15, 16, 17$ changed abruptly, while the remaining ones stay (almost) unchanged. This matches our ground truth data where we manually inject the changes into nodes $13 - 17$.

[4]*GHRG-GLR* is a heurastic-based model, so we dont' compute the likelihood for data fitting

[5]We provide additional results in the supplemental section.

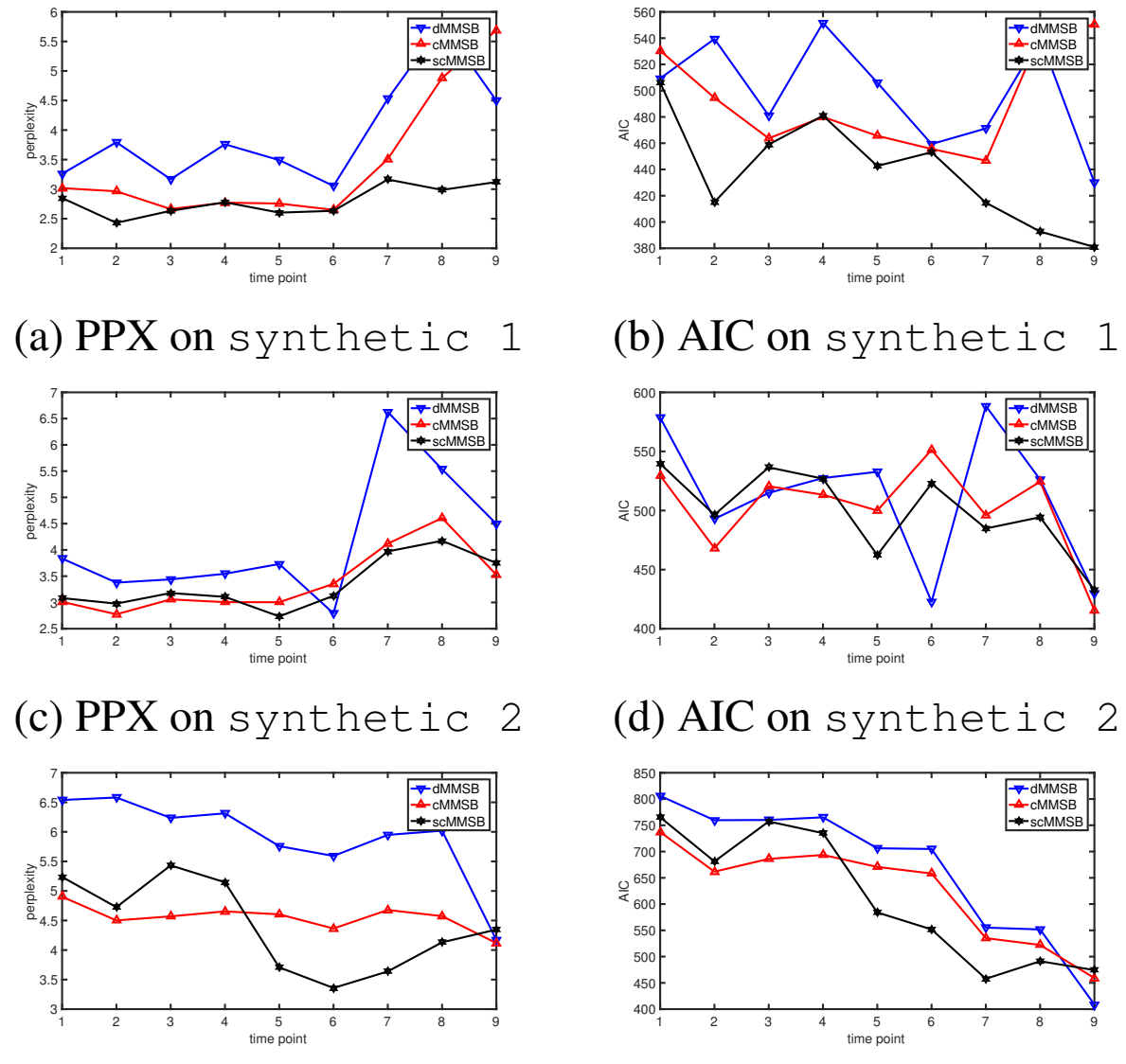

(a) PPX on `synthetic 1` (b) AIC on `synthetic 1`

(c) PPX on `synthetic 2` (d) AIC on `synthetic 2`

(e) PPX on `synthetic 3` (f) AIC on `synthetic 3`

Figure 2: Quantitative evaluation for different models on synthetic data. The performance is measured by PPX (perplexity) and AIC score, which is recorded for each time point.

For *synthetic 3* data, we manually introduce both global and local changes. Figure 4 shows the learned affinity matrix for different models.[6] As we can see, the SC-MMSB model is able to detect two major global changes, and the learned affinity matrix closely matches the ground truth—although the label switching problem still exists. One major difficulty in simultaneously detecting both local and global changes is that the global changes can be absorbed into changes in the node membership. Thus, without necessary conditions such as sparsity constraints, the difference between local and global changes may not be distinguishable.

**Scalability** The major complexity of network learning comes from the growing size of edges; i.e., as the size of $|\mathcal{V}|$ increases, the size of $|E|$ increases quadratically. Under the SGLD framework, we can use a mini-batch for generating samples instead of the full batch of data. Figure 5 shows the comparison results between batch and mini-batch (SGLD) inference algorithms, for different number of nodes. As we can see, both algorithms achieves similar performance but SGLD runs significantly faster.

[6]For brevity, we omit the plots of learned membership vectors.

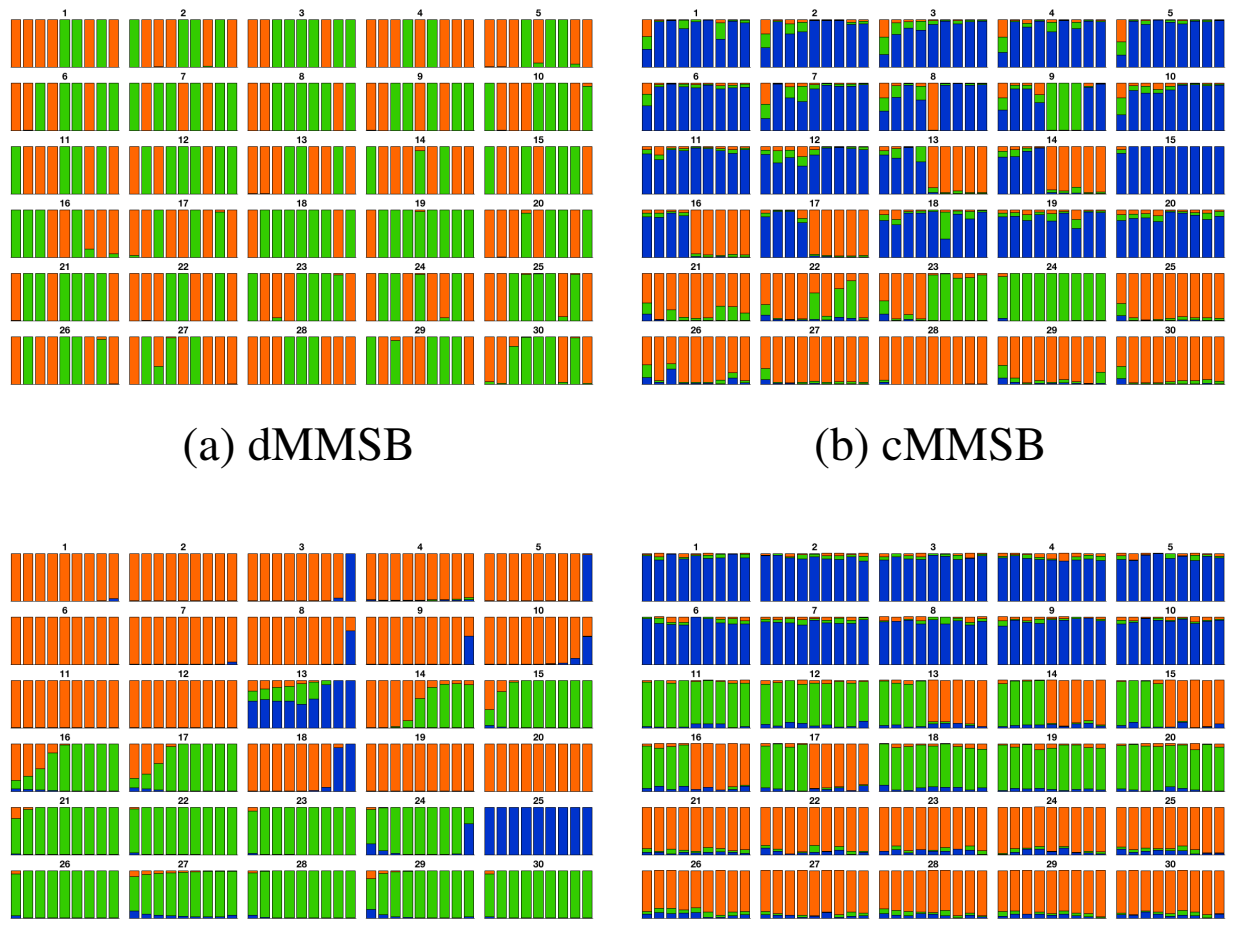

(a) dMMSB (b) cMMSB

(c) SC-MMSB (d) Ground Truth

Figure 3: Learned membership vectors on `synthetic 2` data sets.

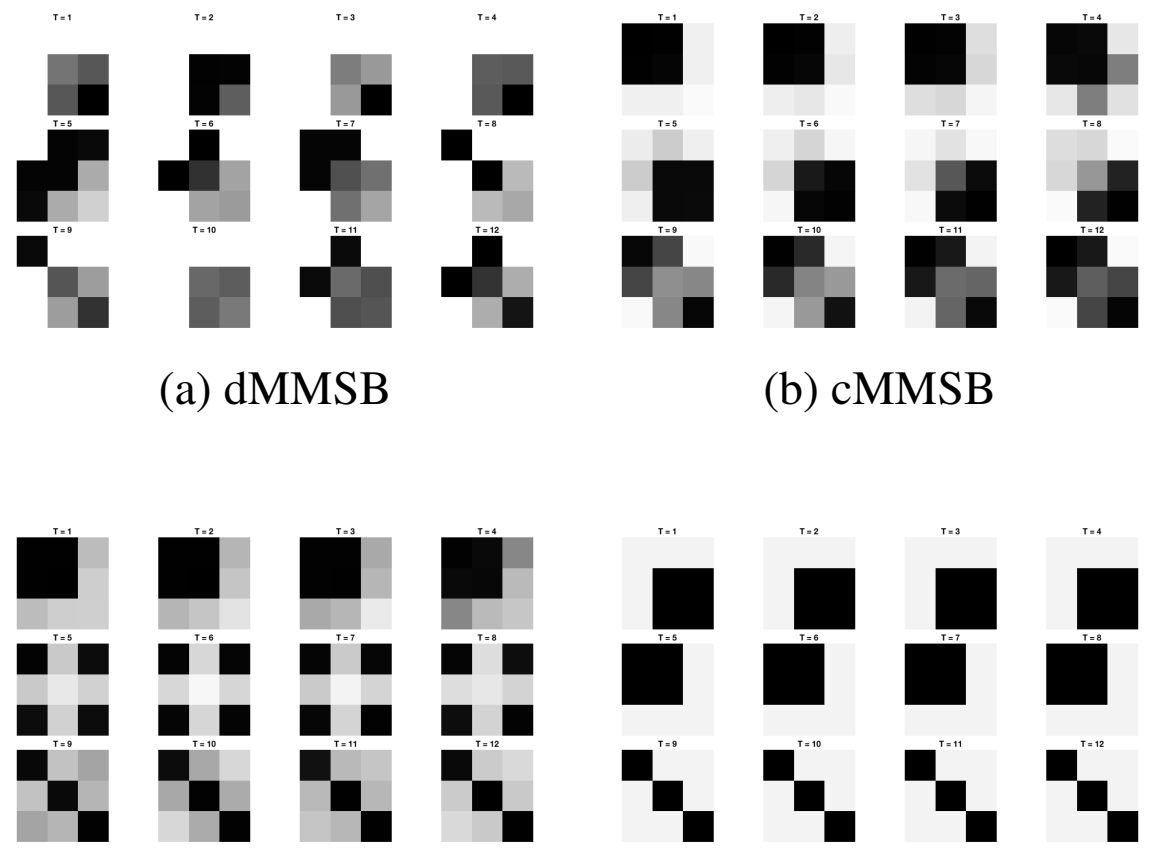

(a) dMMSB (b) cMMSB

(c) SC-MMSB (d) Ground Truth

Figure 4: The learned affinity matrix for different models on *synthetic 3* data.

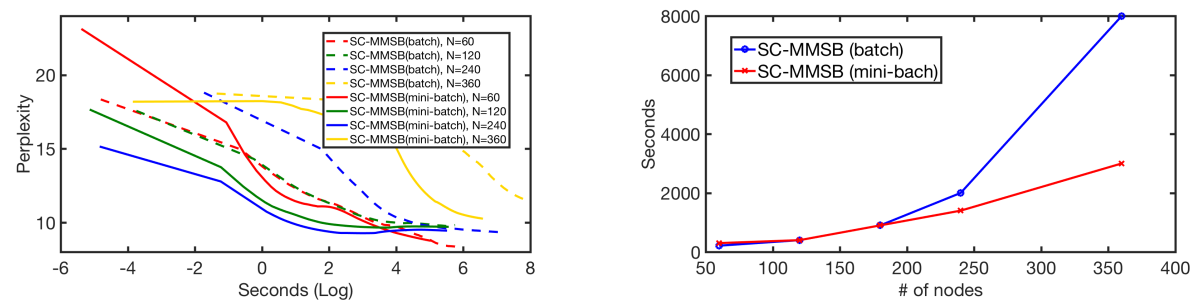

Figure 5: The left figure shows the change of perplexity over time, and the right figure shows the time needed for convergence.

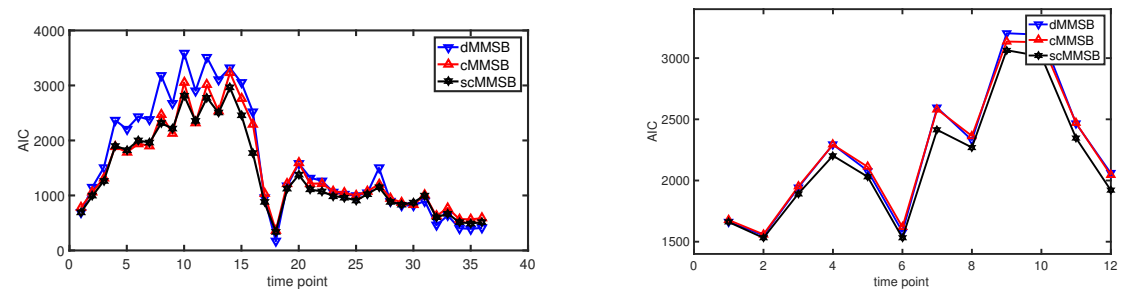

(a) `proximity network` (b) `enron email`

Figure 6: Comparisons of three different models on `proximity network` and `enron email` data sets. The performance is measured by AIC score.

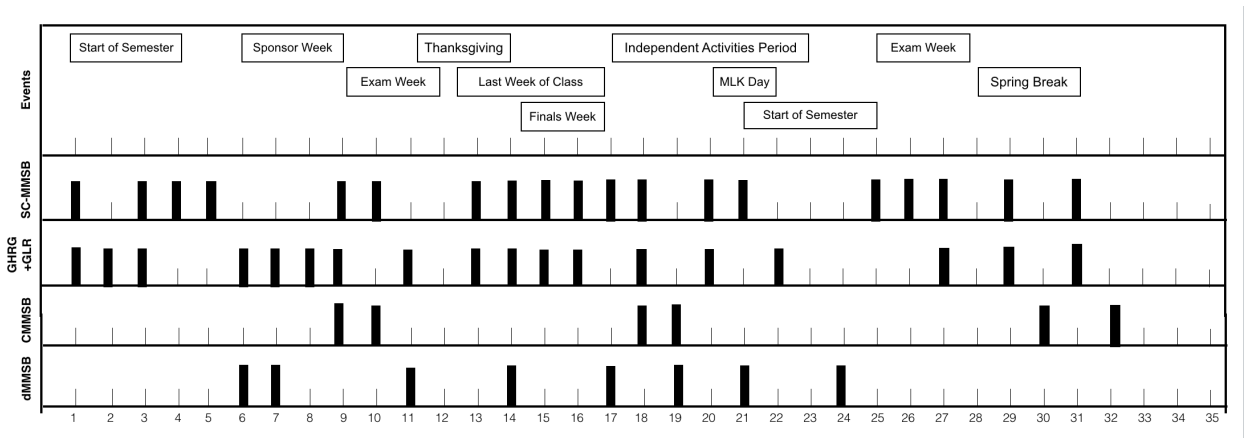

Figure 7: Change points detected for `proximity network`. There are 35 snapshots in total. Each bold bar denotes the detected change point. There are 11 ground truth events (along with duration period) depicted in the first row.

## Experiments on two real-world data sets

In this section we evaluate our model on two real-world data sets, namely, `social proximity network` (Peel and Clauset 2015; Eagle and Pentland 2006) and `enron email` data (Peel and Clauset 2015). Unfortunately, for these two data sets, there is no exact ground truth about the affinity matrix and the local membership vectors for nodes, but the data sets exhibits both local and global changes.

The *social proximity network* consists of 97 faculty and graduate students, whose actions are recorded via mobile phone over 35 weeks. For each week, we extract a snapshot of network from their actions.[7] The edge denotes physical proximity to one of 97 subjects at a given time point. There are 16 known external events such as *public holidays* or *spring break* which we treat as ground truth labels.

Another data set is called `enron email` data. This is an email communication network where an edge denotes a communication between two email users. There are 151 persons in total, and we create each network snapshot from monthly communication data during the year 2001. Thus, we end up with a dynamic network consisting of 12 time points.

Figure 6 shows the AIC score for different models on the two real-world data sets. Similar to the results for synthetic data, our new model SC-MMSB achieves the lowest AIC score on both of the data sets.

We next evaluate the detectability of events on *social proximity network*, which is shown in Figure 7. The ground truth events are obtained from (Peel and Clauset 2015). The discovery of change point is rather simple: After we learn for each model (except GHRG-GLR), we first get a sequence of learned affinity matrices. Then we compute the distance between each consecutive affinity matrix followed by simple thresholding, which results in a series of detected change points. There are a few observations we can make from the results of comparing to dMMSB and cMMSB: (i) Our new model has better accuracy for detecting the correct change points. SC-MMSB is able to detect some major events such as *start of semester, exam week, finals week, MLK Day, start of spring break* etc. On the other hand, cMMSB marks several change-points that are far from any event onset. (ii) Our new model detects major change points more quickly, e.g., using the SC-MMSB model, we are able to detect *finals week* event faster than dMMSB, and discover the *spring break* event more quickly than the cMMSB model. Last, let's compare our model to GHRG-GLR. In terms of detected change points, one major difference is that GHRG-GLR recognizes the sponsor week as a series of major events, while our model does not. This might be due to the sensitivity of the thresholding techniques. But besides this, most of the change points detected by these two methods overlap, providing relatively similar detection performance. However, one thing we should note is that GHRG-GLR is designed for detection of global changes only, while our method is also sensitive to the detection of local changes.

[7] We thank Leto Peel (Peel and Clauset 2015) for kindly providing the code for processing the data.

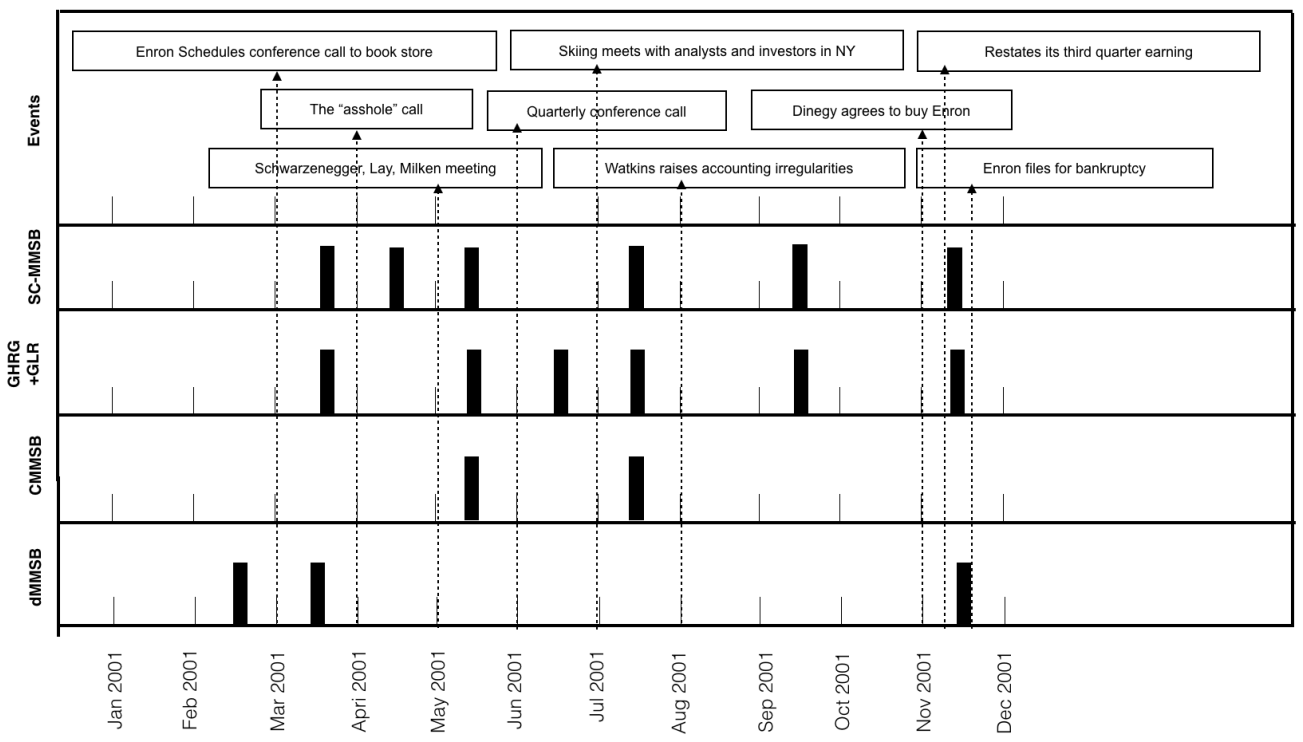

Figure 8: Change points detected for `enron email`. There are 12 snapshots in total. We compare our model with three other baselines. Each bold bar denotes the detected change point. There are 9 ground truth events depicted in the first row.

Similarly, Figure 8 shows the detected change points for *enron email* data. The observations are similar to the results of *proximity network*: SC-MMSB and GHRG-GLR perform noticeably better than the two other baseline methods.

Last, we visualize the learned membership vectors for `enron email` data, as in Figure 9. Because of space limitations, we only show the results for *SC-MMSB*. Here, we set $K = 3$, and there are a total of 151 nodes in the network. We are mainly interested in the nodes that either change abruptly or remain stable throughout the time points.[8] From a quick scan, we notice that the membership for node 31 (Dutch Quigley) changes abruptly during the first half of the time points. After searching external sources, we found that he was in the energy department, and changed his role from risk/legal to full energy traders. It is also similar for node

[8]Since there is no ground truth about node information, we have to search on other resources such as Linkedin.

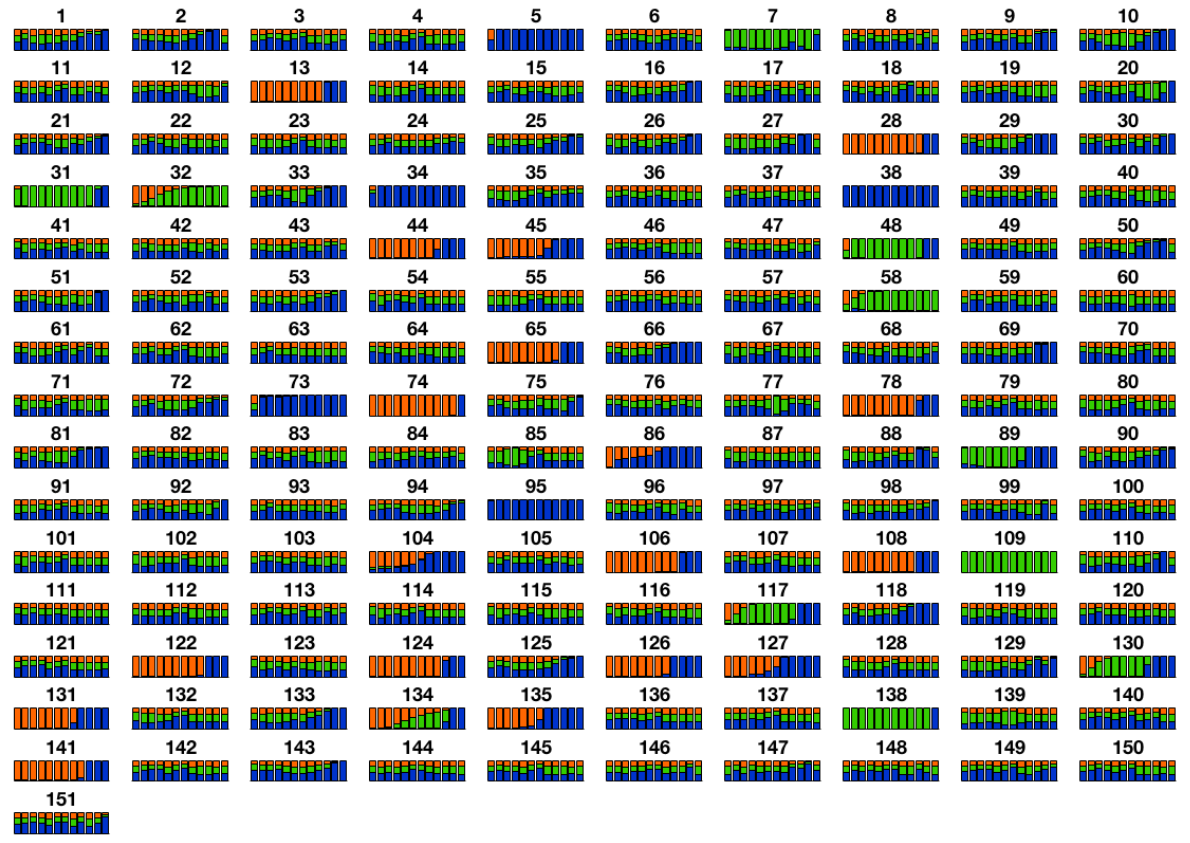

Figure 9: Visualization of learned mixed memberships on `enron email` data from SC-MMSB model. The data set contains 151 nodes and 12 snapshots. We set $K = 3$

95 (Mark Guzman). On the other hand, the membership for node 32 (Mark Haedicke) remains stable, and after doing a quick search, we see that he remains the head of the legal department. The node 104 (Michelle Cash) is another node whose membership remains stable. After searching, we found that she was an assistant general counsel mainly responsible for all aspects of labor and employment law.

## Conclusion and Future Work

In this paper we proposed a novel network model that simultaneously accounts for both local and global dynamics, and provide an efficient inference technique based on stochastic gradient Langevin dynamics (SGLD). Our model is built upon the popular mixed membership stochastic blockmodels with sparse co-evolving patterns. The experiments with both synthetic and real-world data demonstrate that our model performs significantly better than baselines in terms of both quantitative and qualitative measures.

There are several exciting directions for future work. We intend to extend the proposed change detection framework to multilayer networks (Boccaletti et al. 2014), where the changes might affect both intra- and inter-layer connectivity patterns. It will also be interesting to consider networks where in addition to purely local and global dynamics, there are relevant intermediate scales as well, e.g., motifs.

## References

Ahn, S.; Shahbaba, B.; Welling, M.; et al. 2014. Distributed stochastic gradient mcmc. In *ICML*, 1044–1052.

Airoldi, E. M.; Blei, D. M.; Fienberg, S. E.; and Xing, E. P. 2008. Mixed membership stochastic blockmodels. *Journal of Machine Learning Research* 9(Sep):1981–2014.

Akoglu, L.; Tong, H.; and Koutra, D. 2015. Graph based anomaly detection and description: a survey. *Data Mining and Knowledge Discovery* 29(3):626–688.

Bader, G. D., and Hogue, C. W. 2003. An automated method for finding molecular complexes in large protein interaction networks. *BMC bioinformatics* 4(1):2.

Barnett, I., and Onnela, J.-P. 2016. Change point detection in correlation networks. *Scientific Reports* 6:18893 EP –.

Bhadury, A.; Chen, J.; Zhu, J.; and Liu, S. 2016. Scaling up dynamic topic models. In *Proceedings of the 25th International Conference on World Wide Web*, 381–390. International World Wide Web Conferences Steering Committee.

Blei, D. M.; Ng, A. Y.; and Jordan, M. I. 2003. Latent dirichlet allocation. *Journal of machine Learning research* 3(Jan):993–1022.

Boccaletti, S.; Bianconi, G.; Criado, R.; del Genio, C.; Gmez-Gardees, J.; Romance, M.; Sendia-Nadal, I.; Wang, Z.; and Zanin, M. 2014. The structure and dynamics of multilayer networks. *Physics Reports* 544(1):1 – 122. The structure and dynamics of multilayer networks.

Brohee, S., and Van Helden, J. 2006. Evaluation of clustering algorithms for protein-protein interaction networks. *BMC bioinformatics* 7(1):488.

Cadena, J.; Vullikanti, A. K.; and Aggarwal, C. C. 2016. On dense subgraphs in signed network streams. In *2016 IEEE 16th International Conference on Data Mining (ICDM)*, 51–60.

Chiu, C.; Ku, Y.; Lie, T.; and Chen, Y. 2011. Internet auction fraud detection using social network analysis and classification tree approaches. *International Journal of Electronic Commerce* 15(3):123–147.

Cho, Y.-S.; Steeg, G. V.; and Galstyan, A. 2011. Co-evolution of selection and influence in social networks. In *Proceedings of the Twenty-Fifth AAAI Conference on Artificial Intelligence*, AAAI'11, 779–784. AAAI Press.

Eagle, N., and Pentland, A. S. 2006. Reality mining: sensing complex social systems. *Personal and ubiquitous computing* 10(4):255–268.

El-Helw, I.; Hofman, R.; Li, W.; Ahn, S.; Welling, M.; and Bal, H. 2016. Scalable overlapping community detection. In *Parallel and Distributed Processing Symposium Workshops, 2016 IEEE International*, 1463–1472. IEEE.

Fu, W.; Song, L.; and Xing, E. P. 2009. Dynamic mixed membership blockmodel for evolving networks. In *NIPS*, 329–336. ACM.

Gopalan, P. K.; Gerrish, S.; Freedman, M.; Blei, D. M.; and Mimno, D. M. 2012. Scalable inference of overlapping communities. In *NIPS*, 2249–2257.

Grover, A., and Leskovec, J. 2016. node2vec: Scalable feature learning for networks. In *Proceedings of the 22nd ACM SIGKDD International Conference on Knowledge Discovery and Data Mining*, 855–864. ACM.

Hernández-Lobato, J. M., and Adams, R. 2015. Probabilistic backpropagation for scalable learning of bayesian neural networks. In *ICML*, 1861–1869.

Ishwaran, H., and James, L. F. 2001. Gibbs sampling methods for stick-breaking priors. *Journal of the American Statistical Association* 96(453):161–173.

Karrer, B., and Newman, M. E. 2011. Stochastic blockmodels and community structure in networks. *Physical Review E* 83(1):016107.

Kitagawa, G. 1996. Monte carlo filter and smoother for non-gaussian nonlinear state space models. *Journal of computational and graphical statistics* 5(1):1–25.

Kolar, M., and Xing, E. P. 2012. Estimating networks with jumps. *Electron. J. Statist.* 6:2069–2106.

Li, W.; Ahn, S.; Welling, M.; et al. 2016. Scalable mcmc for mixed membership stochastic blockmodels. In *Proceedings of The 19th International Conference on Artificial Intelligence and Statistics*.

Liben-Nowell, D., and Kleinberg, J. 2007. The link-prediction problem for social networks. *journal of the Association for Information Science and Technology* 58(7):1019–1031.

Patterson, S., and Teh, Y. W. 2013. Stochastic gradient riemannian langevin dynamics on the probability simplex. In *Advances in Neural Information Processing Systems*, 3102–3110.

Peel, L., and Clauset, A. 2015. Detecting change points in the large-scale structure of evolving networks. In *Twenty-Ninth AAAI Conference on Artificial Intelligence*.

Raghavan, V.; Steeg, G. V.; Galstyan, A.; and Tartakovsky, A. G. 2014. Modeling temporal activity patterns in dynamic social networks. *IEEE Transactions on Computational Social Systems* 1(1):89–107.

Ridder, S. D.; Vandermarliere, B.; and Ryckebusch, J. 2016. Detection and localization of change points in temporal networks with the aid of stochastic block models. *Journal of Statistical Mechanics: Theory and Experiment* 2016(11):113302.

Roesser, R. 1975. A discrete state-space model for linear image processing. *IEEE Transactions on Automatic Control* 20(1):1–10.

Sakamoto, Y.; Ishiguro, M.; and Kitagawa, G. 1986. Akaike information criterion statistics. *Dordrecht, The Netherlands: D. Reidel*.

Shen, J.; Zhang, J.; Luo, X.; Zhu, W.; Yu, K.; Chen, K.; Li, Y.; and Jiang, H. 2007. Predicting protein–protein interactions based only on sequences information. *Proceedings of the National Academy of Sciences* 104(11):4337–4341.

Šubelj, L.; Furlan, Š.; and Bajec, M. 2011. An expert system for detecting automobile insurance fraud using social network analysis. *Expert Systems with Applications* 38(1):1039–1052.

Wang, Y. J., and Wong, G. Y. 1987. Stochastic blockmodels for directed graphs. *Journal of the American Statistical Association* 82(397):8–19.

Welling, M., and Teh, Y. W. 2011. Bayesian learning via stochastic gradient langevin dynamics. In *Proceedings of the 28th International Conference on Machine Learning (ICML-11)*, 681–688.

Zhu, L.; Guo, D.; Yin, J.; Steeg, G. V.; and Galstyan, A. 2016. Scalable temporal latent space inference for link prediction in dynamic social networks. *IEEE Transactions on Knowledge and Data Engineering* 28(10):2765–2777.

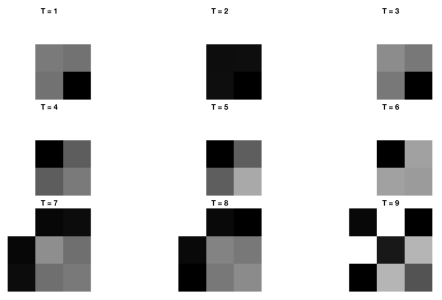

(a) dMMSB

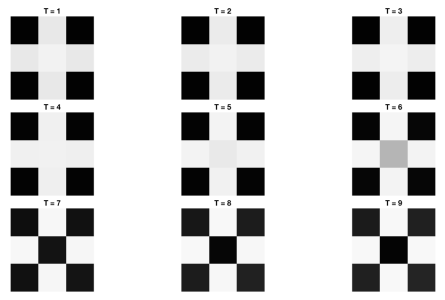

(b) cMMSB

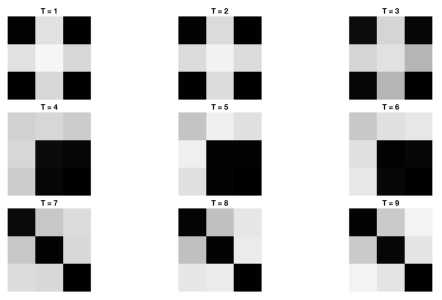

(c) SC-MMSB

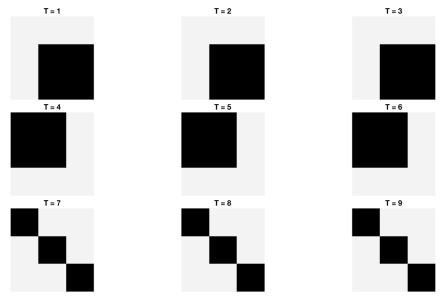

(d) Ground Truth

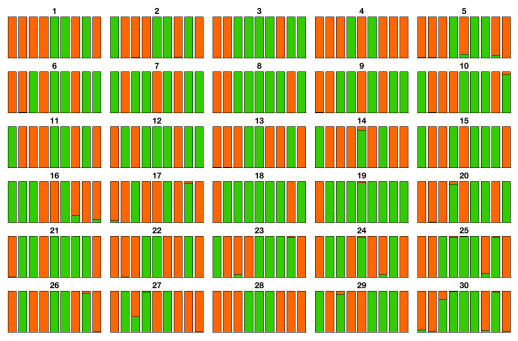

(e) dMMSB

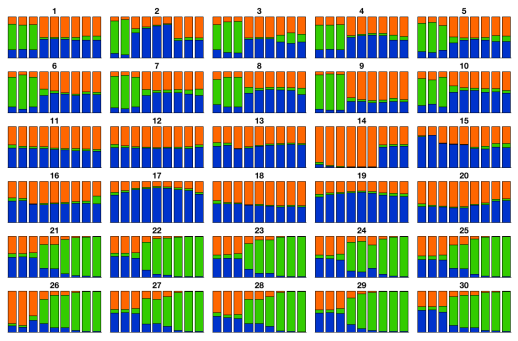

(f) cMMSB

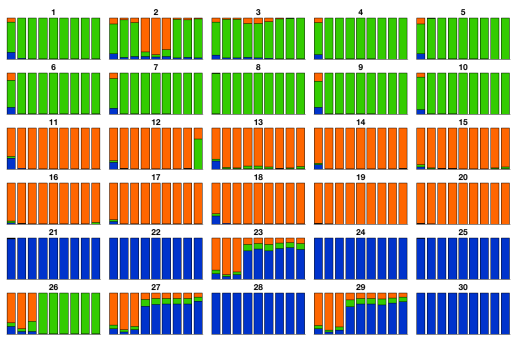

(g) SC-MMSB

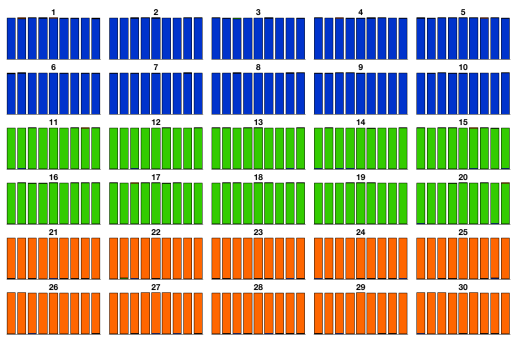

(h) Ground Truth

Figure 10: Learned affinity matrix (figures on the top) and mixed membership vectors (figures on the bottom) on `synthetic 1`. $3 \times 3$ affinity matrix is visualized in black-white color, and the darker region corresponds to larger weight. The size of each segment within the bar chart shows the weight for each community.

## Supplement

In Figure 10 we show the learned affinity matrix and mixed membership vector for each model on *synthetic 1* data. The ground truth is also included. It is clear to see that the cMMSB model with sparse prior (SC-MMSB) performs significantly better than the other two baselines. For the ground truth mixed membership vector, each node belongs to one single community, and SC-MMSB is the only model that can precisely recover this. For the learned affinity matrix, SC-MMSB again does significantly better at recovering the true affinity matrix as well as providing clear change points. On the other hand, the dMMSB model is unable to recover the changing dynamics of the affinity matrix. One thing we should note is that the current model still has the label-switching problem. Because of that, the learned affinity matrix and local membership vectors might be different in terms of their community ordering, which can be seen from the results of SC-MMSB.